\documentclass[runningheads]{llncs}
\usepackage{verbatim}

\usepackage[T1]{fontenc}
\usepackage{graphicx}
\PassOptionsToPackage{hyphens}{url}\usepackage{hyperref}
\usepackage{color}

\usepackage{enumitem}
\usepackage{subcaption}
\usepackage{float}
\usepackage{amsfonts}
\usepackage{algorithmic}
\usepackage{todonotes}
\usepackage[normalem]{ulem}
\usepackage{textcomp}
\usepackage{xcolor}
\usepackage{multirow}
\usepackage[ruled,vlined,linesnumbered]{algorithm2e}

\SetCommentSty{mycommfont}
\usepackage{amsmath}
\usepackage[skip=0pt, indent=15pt]{parskip}
\usepackage[htt]{hyphenat}
\usepackage{array}
\usepackage{caption}
\usepackage{booktabs}
\usepackage{rotating}
\usepackage{adjustbox}
\usepackage{makecell}

\begin{document}

\title{ClustEm4Ano: Clustering Text Embeddings of Nominal Textual Attributes for Microdata Anonymization}
\titlerunning{ClustEm4Ano}
\author{Robert Aufschläger\orcidID{0009-0004-0986-3504} \and
Sebastian Wilhelm\orcidID{0000-0002-4370-9234} \and
Michael Heigl\orcidID{0000-0001-7303-113X} \and
Martin Schramm\orcidID{0000-0001-6206-2969}
}
\authorrunning{R. Aufschläger et al.}
%
\institute{Deggendorf Institute of Technology, Dieter-Görlitz-Platz 1, 94469 Deggendorf, Germany\\
\email{robert.aufschlaeger@th-deg.de}
}

\maketitle              
\begin{abstract}
This work introduces \texttt{ClustEm4Ano}, an anonymization pipe-line that can be used for generalization and suppression-based anonym-ization of nominal textual tabular data. It automatically generates value generalization hierarchies (VGHs) that, in turn, can be used to generalize attributes in quasi-identifiers. The pipeline leverages embeddings to generate semantically close value generalizations through iterative clustering. We applied KMeans and Hierarchical Agglomerative Clustering on $13$ different predefined text embeddings (both open and closed-source (via APIs)). Our approach is experimentally tested on a well-known benchmark dataset for anonymization: The UCI Machine Learning Repository’s Adult dataset. 
\texttt{ClustEm4Ano} supports anonymization procedures by offering more possibilities compared to using arbitrarily chosen VGHs.
Experiments demonstrate that these VGHs can outperform manually constructed ones in terms of downstream efficacy (especially for small $k$-anonymity ($2 \leq k \leq 30$)) and therefore can foster the quality of anonymized datasets. Our implementation is made public.

\end{abstract}
\section{Introduction}

\emph{Background.}
Anonymizing tabular datasets with textual attributes typically requires generalizing textual attributes that qualify as quasi-identifier (QI), i.e., as set of attributes that in combination allow to reidentify individuals \cite{Samarati2001}. For this purpose, VGHs are commonly created to group similar values into categories to prevent privacy attacks, such as record linkage in microdata \cite{Samarati2001}, and to maintain usability. However, creating suitable VGHs is usually a manual process that demands domain knowledge, presenting a challenge for human data processors.

\begin{figure}[h!]
    \centering
    \includegraphics[width=\linewidth]{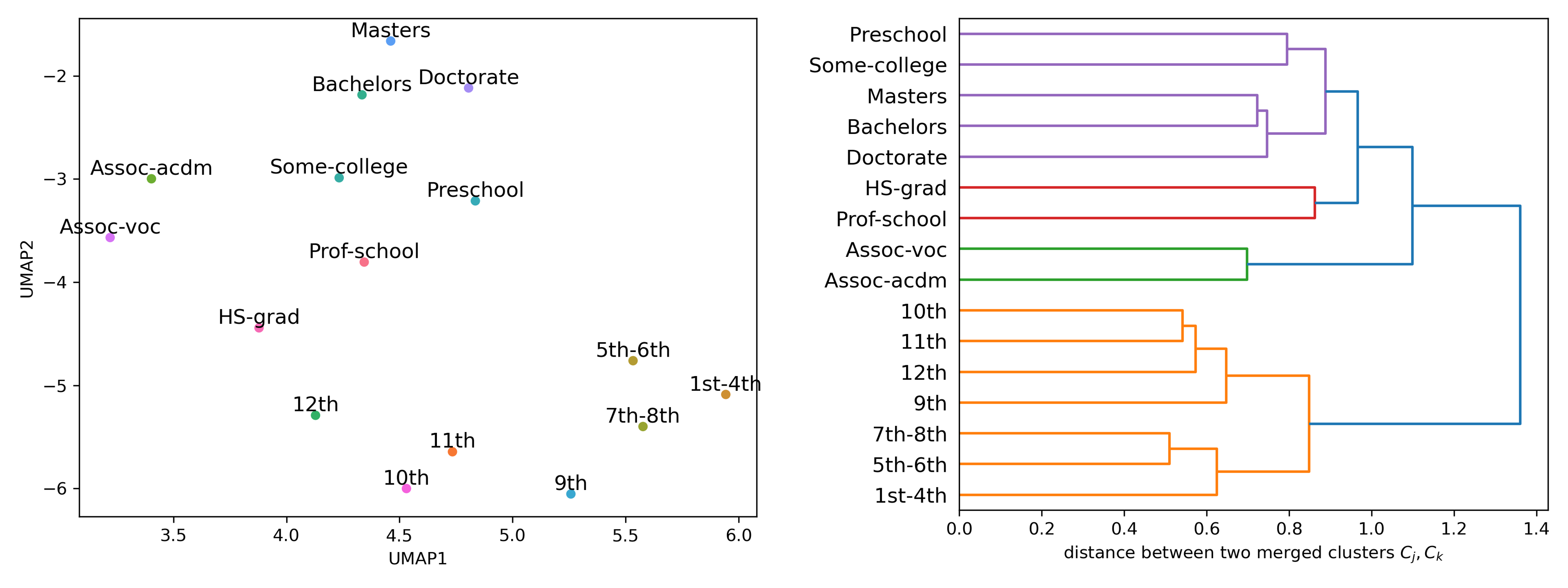}
    \caption{Visualizing \texttt{Mistral AI} embeddings with Uniform Manifold Approximation and Projection (UMAP) (left) and the corresponding VGH (right) obtained
by \texttt{Agglomerative Hierarchical Clustering} of embeddings of the values from the Adult's \cite{misc_adult_2} attribute \texttt{education}.}
    \label{fig:vis-emb}
\end{figure}

\emph{Approach.}
We introduce \texttt{ClustEm4Ano}, a novel approach that employs embeddings to construct VGHs for anonymization. In particular, when considering the uncertainty penalty-based methods to generalize with less information loss, hierarchies of nominal attributes have to be known in advance. This is where text embeddings come in handy due to their intrinsic semantics.
In fact, embeddings can be used to generate VGHs for nominal data that otherwise need to be defined manually or rely on domain orderings like in \cite{Bayardo05}. We experiment with averaging word embeddings and contextualized text embeddings.
We demonstrate that clustering can automate the anonymization of nominal textual data in \emph{microdata}, i.e., tabular data, where each entry is assigned to a single individual. 
Fig. \ref{fig:vis-emb} visualizes the idea of how embeddings relate to clustering and VGH.

\begin{figure}[h!]
    \centering
    \begin{subfigure}{0.49\linewidth}
        \includegraphics[width=\linewidth]{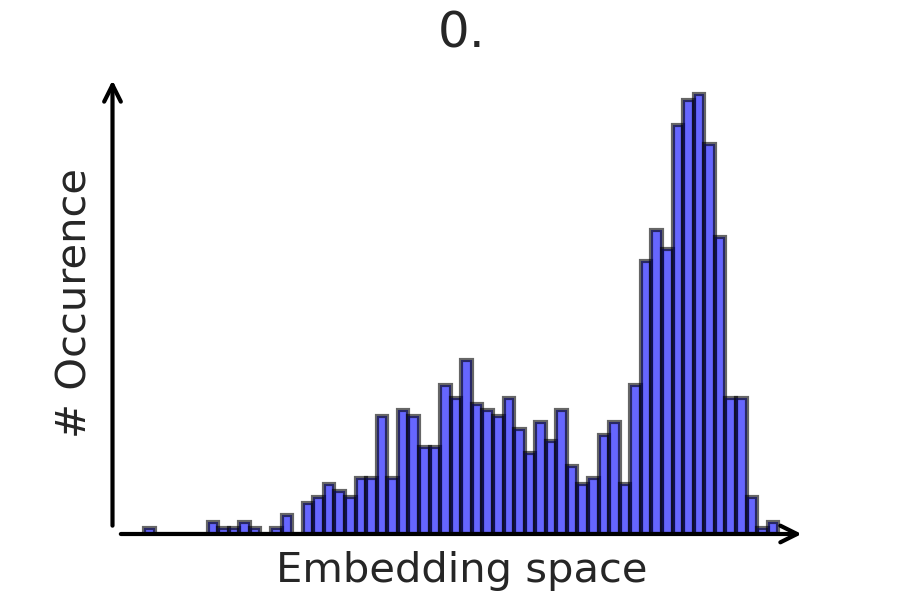}
        \caption{Before anonymization (0.).}
        \label{fig:1a}
    \end{subfigure}
    \hfill
    \begin{subfigure}{0.49\linewidth}
        \includegraphics[width=\linewidth]{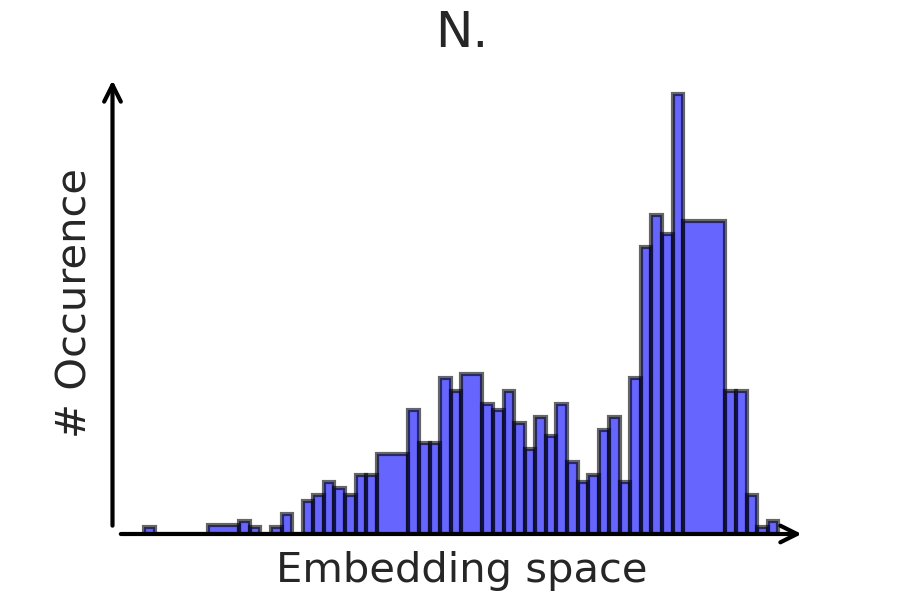}
        \caption{After N generalization steps.}
        \label{fig:1b}
    \end{subfigure}
    \caption{Visualization of the embedding similarity and distribution-based motivation for \texttt{ClustEm4Ano}.}
    \label{fig:intuition}
\end{figure}

A further motivation for the idea to cluster embeddings for anonymization of nominal data is presented in Fig. \ref{fig:intuition}.
In brief, the idea is to use open information about value similarities to maintain data utility while aiming for distribution-invariant anonymization using embeddings. 
In both plots, the $x$-axis represents the similarity between embeddings of nominal attribute values.
The plots would vary for different attributes, and even for the same attribute across different datasets.

We conduct experiments to evaluate the effectiveness of different anonymization strategies on datasets used for training machine learning (ML) models. The experiments involve $13$ different text embeddings and various ML models to assess the classification performance using anonymized data.

\paragraph{Research Questions}
\begin{itemize}
    \item How can semantically accurate VGHs be created using clustering of text embeddings?
    \item How can text embeddings be integrated into existing value generalization and suppression-based anonymization algorithms to meet specific privacy thresholds, such as $k$-anonymity?
    \item How does using different text embeddings affect the VGH-based anonymization process?
\end{itemize}

\paragraph{Contributions}
\begin{itemize}
    \item We introduce an anonymization pipeline \texttt{ClustEm4Ano}, designed to anonymize microdata that have nominal textual quasi-identifier attributes.
    \item As part of that, we propose the use of automatically generated VGHs derived from clustering text embeddings from nominal attribute values.
    \item We experimentally compare the use of different text embeddings to create VGHs for anonymization and compare them to manually defined VGHs (baseline) with promising results in retained percentage (after suppression), normalized average group size metric, and ML efficacy (test accuracy and $F1$ Score after training ML classifiers with anonymized Adult datasets).
\end{itemize}

\emph{Availability.}
Our code is made available as open-source on GitHub at \textcolor{blue}{\url{https://github.com/EAsyAnon/ClustEm4Ano}}.

\section{Related Works}
\label{related_work}

\emph{Text Anonymization.}
Due to the intricate nature of natural language, anonymizing textual data presents a complex challenge \cite{Hassan2019,Medlock2006}. 
While numerous methods for anonymizing structured data with well-established privacy models are available, they are not directly transferable to text data, so particular approaches are necessary \cite{Lison2021}. 
One of the earliest contributions to text data anonymization dates back to \cite{Sweeney1996}, which details the identification and replacement of personally identifying information in medical records using various pattern-matching algorithms.
Over the years, various strategies such as removal, tagging, generalization, and substitution have been developed.
The choice of strategy depends on the context of the application and varies significantly between anonymizing free-form text, like letters, and structured categorical data, like microdata. 

\emph{VGHs for Anonymization of Microdata.}
The generalization process of textual attributes for microdata anonymization relies primarily on a specific VGH \cite{Samarati2001}, typically obtained from a knowledge base \cite{Mamede2016,AyalaRivera2017}, and the algorithms to obtain $k$-anonymity are based on a subsequent generalization lattice for the quasi-identifier attributes \cite{ELEmam09}. The creation and influence of such VGHs have been extensively discussed, but existing research on creating VGHs either uses specific knowledge bases, as in \cite{AyalaRivera2016}, focuses on numerical data \cite{Campan2011}, or uses semantic rules obtained from human expert knowledge \cite{Mubark2016}.

\emph{NER-Based Anonymization for Textual Data.}
Named Entity Recognition (NER) is widely used for identifying sensitive entities, including categories such as individuals, organizations, and geographic locations in free-form text. NER employs both rule-based systems and ML techniques, such as Hidden Markov Models and Conditional Random Fields \cite{Wellner07}. Post-processing often includes contextual analysis to identify additional sensitive data \cite{Mamede2016}. However, these methods are less effective for structured data and often lead to significant information loss or distortion \cite{Hassan2019,Lison2021,Biesner2022}. The methods also depend on extensive volumes of annotated datasets, particularly data that has already been correctly anonymized.

\emph{Using Text Embeddings for Anonymization.}
Word embeddings represent words in a high-dimensional space, capturing semantic relationships, which are useful in NER and potentially for anonymization \cite{Hassan2019,Biesner2022,Abdalla2020}. Text embedding models are typically learned on extensive domain-specific text corpora and primarily rely on neural networks. 
The vectorization process results in a token representation that is subsequently mapped to probabilities for specific entities -- for example, to rate sensitivity \cite{Biesner2022}. 
Nevertheless, this approach also relies on the definition of subjects to be protected or on expert-annotated data, but in much smaller amounts than for NER-based technologies \cite{Hassan2019}.
In \cite{Hassan2019}, entities in text documents are represented as word vectors that capture semantic relationship. Then sensitive entitities can be protected by removing other entities co-occuring, where the word vectors are similar. In \cite{Abdalla2020}, tokens are replaced with tokens that have neigboring word vectors in the embedding space, and in \cite{Biesner2022}, token embeddings are classified and sensitive tokens are replaced by class tokens.

\emph{Using Clustering for Anonymization.}
Clustering methods have been applied to various data types, including text. 
Referring to free-text data, in \cite{Mehta2021}, the authors use word embeddings from \emph{BERT} and propose a clustering method for large text documents.
Another example of clustering-based anonymization of text data is provided in \cite{Garat2022}, where the authors fine-tuned a NER tool and solved co-references between named entities by clustering. 
Further, a relevant work is \cite{Byun07}, where the authors propose a greedy $k$-member clustering algorithm that adds records to a randomly chosen record individually while always adding the candidate record with the smallest information loss in the cluster after adding. This process is repeated while always using a record whose distance is most far away from the previous starting record.
In \cite{Lin08}, the authors use \texttt{KMeans} clustering to simultaneously partition records into clusters with similar records and then adjust the group members to obtain $k$-anonymity. 

\emph{Placement of This Work.}
Our analysis identifies a gap in the literature concerning the automatic generation of VGHs using word embeddings and clustering.
Current methods mainly focus on NER systems and do not fully utilize the potential of VGHs. 

\section{Basic Definitions}
\label{definitions}

\emph{Anonymization.}
Let $D=\{R_1, R_2, \ldots, R_n\}$ be a multiset of row records with $n \in \mathbb{N}^+$ not necessarily distinct records composed of attribute values.
We refer to an \emph{attribute} as a peace of information about an individual.
A QI is a set of attributes that allows re-identification of individuals in data records.
\emph{Sensitive Attributes} (SAs) are attributes in $T$, where it should be impossible to assign corresponding attribute values after potential attribute inference (attacks). 
In \emph{domain} generalization \cite{Samarati2001}, the domains of QI's attributes are mapped to a more general domain. Domain Generalization can be achieved via different VGHs, where each value in a non-generalized domain is mapped to a unique value in a more general domain.
If all generalized values are  on the same level in the value generalization hierarchy, it is called \emph{full-domain} generalization.
In general, VGHs are a subclass of ontologies where only the is-a (sub-type) relationships are considered \cite{AyalaRivera2017}.
In particular, in our experiments, we use the generalization algorithm \texttt{FLASH} \cite{Kohlmayer2012}, which, among others, is implemented by the ARX Software~\cite{Prasser2014}. This algorithm offers fast execution times and seeks minimal information loss while traversing the generalization lattice in a bottom-up breadth-first search. Also, the anonymization algorithm allows to predefine certain required thresholds for \emph{privacy models} such as \emph{$k$-anonymity} \cite{Samarati2001,samarati1998protecting}, \emph{$l$-diversity} \cite{machanavajjhala2006l}, and \emph{$t$-closeness} \cite{li2006t}.
Thereby, records with identical QI attributes are summarized into a single equivalence class $group$. We denote the set of all equivalence classes as $groups$. Then, the privacy models are defined as follows: $k$-anonymity (\(k\)): $k := \min_{group \in groups} \lvert group\rvert$, $l$-diversity (\(l\)): $l := \min_{group \in groups} |\{\text{values for SA in $group$}\}|
$, $t$-closeness (\(t\)): checks granularity of SA values in single groups in comparison to the distribution in the dataset.
Each of the privacy models has its advantages and limitations \cite{Aufschlaeger2023}: 
$k$-anonymity effectively prevents singling out individuals but depends on $k$. $l$-diversity improves on $k$-anonymity by adding protection against inference. $t$-closeness provides additional defense against inference by ensuring that the distribution of sensitive attributes in each group closely matches the overall distribution.

\emph{Word embeddings.}
Dealing with nominal textual attributes we transform text into numerical representations. A distributed word representation that is dense, low dimensional, and real-valued, is called \emph{word embedding} \cite{Turian2010}.

\emph{Clustering.}
\emph{Clustering} is an unsupervised ML method that takes as input $n$-dimensional real-valued feature vectors, where $n \in \mathbb{N}$, $K \in \mathbb{N}$ the number of desired cluster centers, and outputs $n$-dimensional real-valued cluster centers $c_1, c_2, \ldots, c_K$ and assigns each feature vector exactly one cluster center. We utilize two bottom-up clustering techniques. First, a clustering sequence that incorporates the non-hierarchical clustering method \texttt{KMeans} \cite{Lloyd1982} and, second, Ward's \texttt{Hierarchical Agglomerative Clustering} \cite{ward1963hierarchical}, where two clusters are successively merged and the variance of the clusters being merged is minimized. While \texttt{KMeans} is not designed for hierarchical clustering, Ward's \texttt{Hierarchical Agglomerative Clustering} has a hierarchical structure, and among the Agglomerative Clustering approaches, Ward provides the most regular sizes\footnote{\url{https://scikit-learn.org/stable/modules/clustering.html\#hierarchical-clustering}, accessed on 13/02/2024}, i.e., most similar cluster sizes.  It is expected that \texttt{KMeans}' properties are beneficial: Separates samples in K groups (allows to specify number of groups, which is the only input we need), creates disjoint clusters, clusters are chosen to have similar variance, scales well to large number of samples, is widely used in many different fields.
However, on the other hand, \texttt{KMeans} minimizes so-called inertia, which assumes clusters in the embedding space to be isotropic and convex.
When it comes to isotropy in the encoding clusters, the authors in \cite{cai2021isotropy} found that \emph{BERT} has a smaller average Local Intrinsic Dimension (LID) than larger language models such as \emph{GPT} and \emph{GPT2} in almost all hidden layers (resp. contextual embeddings). Thereby, smaller LIDs indicate local anisotropy that potentially could lead to a less qualitative clustering. We do not pursue further research in this direction.

\section{Proposed Pipeline}
\label{method}

\begin{figure}[h]
    \centering
    \includegraphics[width=0.9\linewidth]{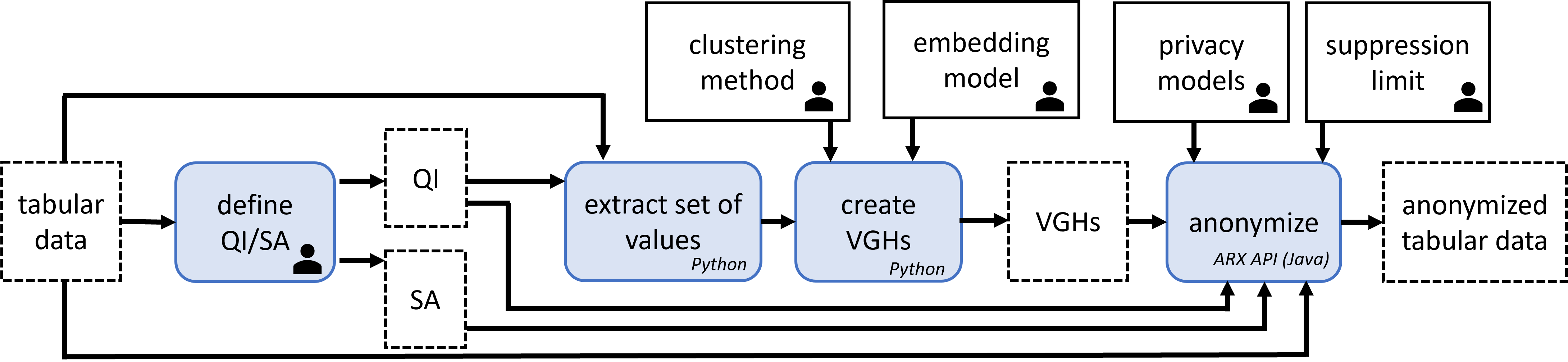}
    \caption{\texttt{ClustEm4Ano}: Dataflow diagram. The human piktogram denotes variables and processes that need to be given by a user.}
    \label{fig:overall_system}
\end{figure}

Fig. \ref{fig:overall_system} shows the dataflow diagram of \texttt{ClustEm4Ano}. First, a data processor (denoted by the human symbol) needs to define a set of QIs with at least one entry and optionally a sensitive attribute. Then, the attribute values from the QI(s) column(s) are extracted. Given the set of extracted values, a predefined clustering method (\texttt{Hierarchical Agglomerative Clustering} / iterative applying \texttt{KMeans}), and an embedding model (for example, one of the $13$ tested experimentally in this pipeline), can be used to generate VGH(s).
Using the generated VGHs, predefined privacy models like $k$-anonymity, and a suppression limit (which defines the maximum percentage of records where all QI attributes are replaced with '*' to handle rare values or outliers), anonymized tabular data can be produced.

We propose the anonymization pipeline in Algorithm \ref{alg:ClustEm4Ano}. One novelty is introduced in Algorithm \ref{alg:createVghForAttribute}, which utilizes clustering of text embeddings to obtain multi-level (i.e., with multiple hierarchies), full-domain generalization (i.e., mapping a chosen value to the same generalized value or value set across all records). 
Our implementation uses \texttt{KMeans} and \texttt{Agglomerative Hierarchical Clustering} to group similar values together based on their distance in the embedding space. \texttt{KMeans} works by partitioning the whole dataset into clusters such that each embedding belongs to the cluster with the nearest center. The proposed algorithm first transforms each unique value in the column to an initial set of raw values and a parallel set of preprocessed values. Initially, each unique value is its own cluster and is associated with a unique numeric label. Over time, in each iteration of the while loop in Algorithm \ref{alg:createVghForAttribute}, the algorithm merges clusters into fewer and fewer clusters. The process is repeated until achieving a fully generalized state, where all values belong to a single cluster -- represented by '*'. When using \texttt{KMeans}, we noted that, when fitting center vectors in early stages in the generalization algorithm, the \texttt{KMeans} implementation in \texttt{sklearn} sporadically did not converge, leading to inconsistent labels and cluster centers. For example, in the experiments, when clustering the \texttt{word2vec} embeddings of the attribute \texttt{native-country} with $42$ values, in the first generalization step $7$ values were generalized in one cluster, and in the $4$ subsequent steps there was no generalization (only from the $5$th step the \texttt{KMeans} algorithm input \texttt{n\_clusters} was small enough that the algorithm could converge). This, however, did not result in worse performance in the experimental results.
Besides \texttt{KMeans}, we applied \texttt{Agglomerative Hierarchical Clustering}, where pairs of clusters are recursively merged based on linkage distance between clusters, not only the distance between the cluster centers. In the anonymization (Algorithm \ref{alg:ClustEm4Ano}), in line $2$, we use the algorithm \texttt{FLASH} \cite{Kohlmayer2012} -- a globally-optimal anonymization algorithm with minimal information loss.

\begin{algorithm}[h]
  \SetKwData{Input}{Input}
  \SetKwData{Output}{Output}
  \SetKwFunction{values}{values}
  \SetKwFunction{preprocess}{preprocess}
  \SetKwFunction{vectorize}{vectorize}
  \SetKwFunction{getCluster}{getCluster}
  \SetKwFunction{getCategories}{getCategories}
  \SetKwInOut{Input}{Input}
  \SetKwInOut{Output}{Output}

    \Input{$T$: Table of records \\ $q$: Attribute column in $T$}
    \Output{vgh: Value generalization hierarchy}

values = [\preprocess{value} for value in \values{$T[q]$}]\;

vgh = [values]\;

labels = [0, ..., num\_of\_values - 1] \;
cluster\_centers = [\vectorize{value} for value in values]\;

\While{\#cluster\_centers > 1}{
    labels(, cluster\_centers) = \getCluster{labels, n\_clusters = len(values) - i, vectors = cluster\_centers}; \tcp{KMeans: labels, cluster\_centers}
    categories = \getCategories{values, labels}\;
    vgh.append(categories)\;
}

\Return vgh\;
  \caption{\texttt{create\_vgh\_for\_attribute}}
  \label{alg:createVghForAttribute}
\end{algorithm}
\vspace{-1.5cm}
\begin{algorithm}[h!]
  \SetKwData{Input}{Input}
  \SetKwData{Output}{Output}
  \SetKwFunction{FLASH}{FLASH}
  \SetKwFunction{preprocess}{preprocess}
  \SetKwFunction{createVghForAttribute}{create\_vgh\_for\_attribute}
  \SetKwInOut{Input}{Input}
  \SetKwInOut{Output}{Output}

  \Input{$T_{original}$: Original table of records \\ $QI=\{q_1, q_2, \ldots, q_m\}$: QI attributes in $T$ \\ 
  $k$: Desired $k$-anonymity \\
  $l$: Desired $l$-diversity \\
  $sup\_lim$: Suppression limit}
  
  \Output{$T^\star$: Table of anonymized records}

  $vghs$ = $\{$\createVghForAttribute($T$,$q_j$) \ : \ $j \in 1..m$$\}$\; 

  $T^\star$ = \FLASH{$T, QI, SA, vghs, k, l, sup\_lim$}; \tcp{anonymization}\ 
  \Return{$T^\star$};

  \caption{\texttt{ClustEm4Ano}}
  \label{alg:ClustEm4Ano}
\end{algorithm}
\vspace{0.1cm}

\section{Experimental evaluation}
\label{experiments}

\subsection{Methodology}

In the context of anonymization, evaluation can be applied in several steps. Before applying anonymization, a domain expert must choose IDs, QIs and SA(s). To the best of our knowledge, there is no automatic evaluation assigning these, and they must be manually defined.
After applying the anonymization algorithm,
we evaluate the data by measuring its efficacy in a downstream classification task. Additionally, we check the number of suppressed records and how close the data matches the desired $k$-anonymity. Also, from a privacy perspective, privacy risks are assessed at this step.
We only perform a quantitative research. 

\subsection{Implementation details}

\emph{Embeddings.}
It can be assumed that there is no ``one [embedding] fits all'' because no particular text embedding method dominates across all downstream tasks, as shown in the MTEB: Massive Text Embedding Benchmark \cite{muennighoff2023mteb}.
Besides using open-source embeddings, we also test the possibility of retrieving embeddings by accessing proprietary APIs. We experiment with OpenAI's text embedding models \texttt{text-embedding-3-small} (\$0.02 / 1M tokens) and \texttt{text-embedding-3-large} (\$0.13 / 1M tokens) 
- released on January 25, 2024 and Mistral AI`s text embedding model \texttt{mistral-embed} (\$0.1 / 1M tokens).
Overall, due to the small number of tokens needed, the price of using OpenAI and Mistral AI embeddings in our experiments was negligible ($<0.1\$$ in total).
We use pre-trained word embedding based on \emph{word2vec} \cite{Mikolov2013} that was trained on the text corpus obtained from \texttt{Google News}.
Thereby, we use the implementation from the \texttt{Python} library \texttt{Gensim} \cite{Rehurek2010}. Further, we test our approach on other pre-trained word embeddings based on \emph{BERT} \cite{Devlin2018,guenther2023jina,reimers-2019-sentence-bert}, \emph{GloVe} \cite{Pennington2014}, \emph{MPNet} \cite{song2020mpnet}, OpenAI's closed-source models \texttt{text-embedding-3-small} and \texttt{text-embedding-3-large}, and \emph{fastText} \cite{grave2018learning}.
We accessed most of the embeddings through Hugging Face's \texttt{SentenceTransformers} \texttt{Python} famework\footnote{\url{https://huggingface.co/sentence-transformers}, accessed on 27/01/2024}.
The complete list of embeddings is given in Section \ref{experiments}. With the embedding models \texttt{word2vec-google-news-300}, \texttt{bert-base-uncased}, and \texttt{fastText/cc.en.300.bin}, we use averaged word embeddings. Otherwise, we use embeddings that transform the whole textual attribute value at once.
Whereas in \emph{word2vec} and \emph{GloVe}, a word is mapped to exactly one embedding vector, making them non-contextual word embeddings, in the embeddings obtained from the transformer-based language model \emph{BERT}, or from OpenAI's embeddings, the context of a word can be used to find a corresponding embedding. Therefore, the same word can lead to different embedding vectors. However, in our experiments when we use pre-trained contextual embeddings, we simplify by considering the single words without contextual information and average their embeddings. Only when we use Mistral AI and OpenAI's embedding models we use text embeddings that take into account the contextual information of the whole textual attribute value.

\emph{Clustering.}
The clustering is implemented in \texttt{scikit-learn} \cite{scikit-learn}.
When using the \texttt{scikit-learn}'s clustering algorithms, we use the default parameters in the constructor except for the number of initial clusters.
For all implementations except for the anonymization algorithm we used \texttt{Python} (3.11.8).

\emph{Anonymization.}
Once given the VGHs we can conduct anonymization through generalization and suppression automatically by using the ARX Software \cite{Prasser2014}. 
We use the anonymization algorithm implemented in \texttt{Java} (java.version=17.0.10) in the ARX Software\footnote{\url{https://arx.deidentifier.org/development/algorithms/}, accessed on 30/01/2024}.
The tool conducts an efficient globally-optimal search algorithm for transforming data with full-domain generalization and record suppression. 
For practical reasons and plausibility, we fix one QI set that implies higher or equal $k$-anonymity for any QI that is a subset of the larger QI. The approach to choose only one (large) quasi-identifier is intrinsically most easy and most secure when it comes to $k$-anonymity. From a technical perspective, we only experiment using textual and numerical attribute values with the data types string and float, respectively. 

\subsection{Setup} 

\emph{Dataset.} In this study, we utilize the Adult dataset \cite{misc_adult_2} and its official train-test split.
In our experiments, we use nominal attributes with expected high feature importance, define QI:=$\{$workclass, education, occupation, native-country$\}$, and the attribute salary-class is the sensitive attribute SA.

\emph{Configuration.}
We use several values $k = 2,5,10,15,20,25,30,50,100,150,200$ for certain $k$-anonymity as input for the anonymization algorithm in our evaluation. We use a suppression limit of $50\%$ which increases speed in anonymization and masks outlier values. Further, we enforce $l$-diversity $\geq 2$. Because there are only two possible values for SA, we obtain $l=2$ in all anonymizations. We do not enforce $t$-closeness because, given only two possible values, the distributions of groups tend to be more concentrated. In such cases, if a group ends up with an overrepresentation of one value, it can significantly skew the distribution away from the dataset’s overall distribution. On the contrary, if a sensitive attribute has many possible values, the distribution in a group can be quite spread out and diverse even if some values are overrepresented.
As a baseline, we use the same VGHs as applied in \cite{Diaz2023}, where the hierarchies are extracted from the official ARX Software GitHub repository.
The baseline VGHs have significantly smaller number of levels in the hierarchies ($3-4-3-3$) compared to our approach where the number of levels equal the number of attribute values (see Algorithm \ref{alg:createVghForAttribute}).
We only find hierarchies for textual attributes.
After anonymization, the order of the rows is not altered.

\emph{Evaluation criteria.}
In our study, we evaluate various measures to assess the trade-off between privacy preservation and utility in the anonymization process. Additionally, we employ several ML models to gauge the impact of anonymization on predictive performance in binary classification. We calculate the measures given the anonymized data $D'$, where suppressed records were removed. For ML efficacy, we include all data, including the suppressed records, where the QI attributes have value '*'. $D$ denotes the un-anonymized dataset. Besides rating privacy preservation in terms of \emph{privacy models} as discussed in Section \ref{definitions}, we evaluate the anonymized dataset in terms of utility and ML efficacy. \emph{Utility criteria}:
Percentage of records retained (\text{perc\_recs}): Ratio of anonymized records to original records, indicating the data suppression: $
perc\_recs := |D'| / |D|$, Normalized average group size metric \cite{lefevre2006mondrian}:
$c_{avg}:= |D'| / (|groups|\cdot k)$.

\emph{Machine learning efficacy.}
In advance, before using the data as input for ML models, we perform (multi-label) one-hot encoding on nominal textual data. We do not create VGHs for numerical data as we focus on textual attribute values in our approach.
We evaluate by training ensemble ML models with anonymized data (features: QI $\cup$ $\{$\texttt{capital-gain}, \texttt{capital-loss}, \texttt{hours-per-week}$\}$, label: SA).
We perform simple hyperparameter tuning in our experiments with \texttt{sklearn's} \texttt{GridSearchCV} with $5$-Fold Cross Validation. We use the classification accuracy ($ACC$) for scoring, and after finding the best hyperparameters, we calculate $ACC$ and $F1$ Score applying the trained binary-classification ML model on the unseen test dataset.
For the Adult dataset, we use the sensitive attribute salary-class as target variable for classification as a downstream ML task. 
We use $F1$ Score, because it helps with high class-imbalance in the salary-class as it combines both precision and recall.
We include the Random Forest classifier (RF) \cite{breiman2001random}, AdaBoost \cite{AdaBoost} with base classifier decision tree, and Bagged Decision Trees (BAG) \cite{breiman1996bagging} in our experiments.
We tuned the RF model using the following hyperparameter grid: [n\_estimators: [10, 50, 100] (number of trees), max\_depth: [None, 10, 20, 30], min\_samples\_split: [2, 5, 10] (min. samples required to split an internal node), min\_samples\_leaf: [1, 2, 4] (min. samples required to be at a leaf node)]. We configured the AdaBoost/BAG models with a \texttt{DecisionTreeClassifier} as \texttt{base\_estimator} and used the following hyperparameter grid for tuning:[\texttt{base\_estimator\_\_criterion}: ['gini', 'entropy'], \texttt{base\_estimator\_\_max\_depth}: [4, 5, 6, 7, 8, 9, 10, 11, 12, 15, 20, 30, 40, 50, 70, 90, 120, 150], \texttt{n\_estimators}: [10, 50]].

\subsection{Experimental results}

\begin{figure*}[h]
    \centering
    \includegraphics[width=\linewidth]{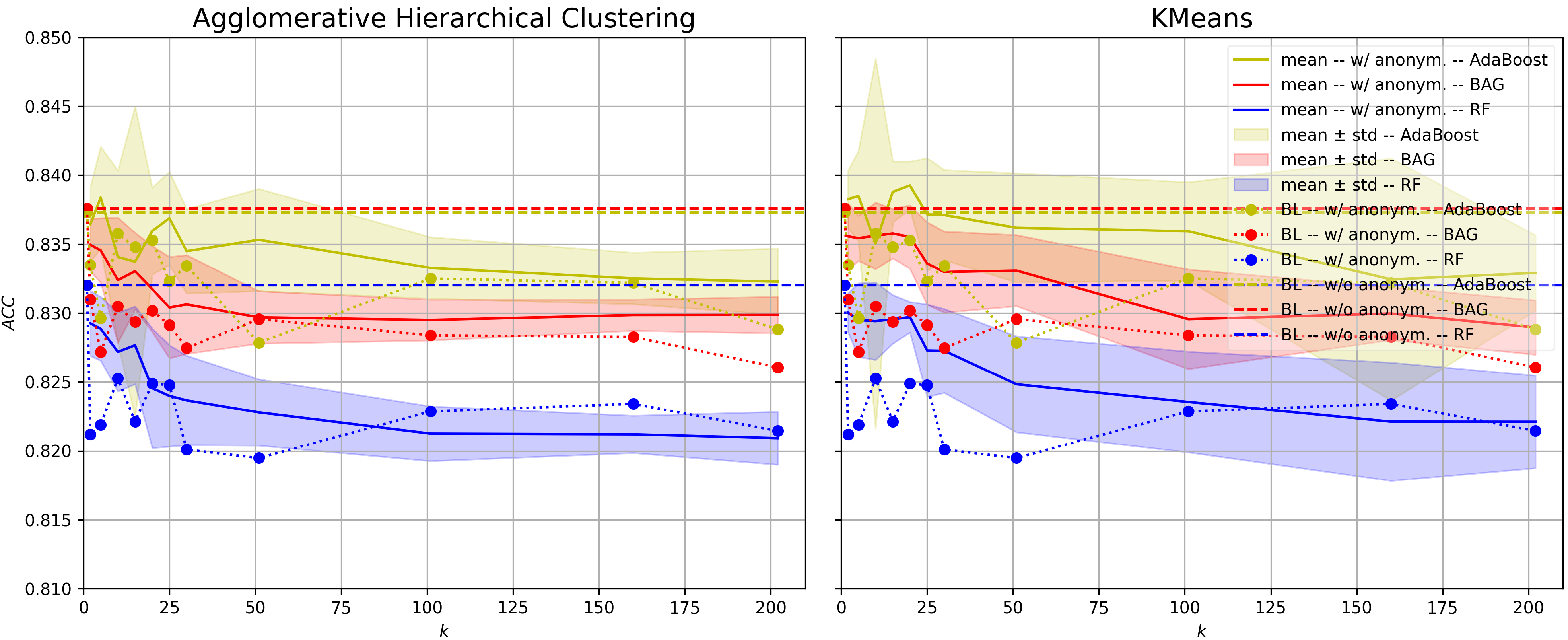}
    \caption{Accuracy comparison between models trained on anonymized data obtained using VGHs obtained by \texttt{Agglomerative Hierarchical Clustering} (left) and \texttt{KMeans} clustering (right).}
    \label{fig:acc-comp}
\end{figure*}

Fig. \ref{fig:acc-comp} compares ML efficacy in terms of $ACC$ using different clustering methods and different ML models in our approach. 
To speed up training and to increase the importance of generalized values in the ML efficacy evaluation, we only use the QI columns and the SA in training and testing.
For $k\leq30$ the \texttt{KMeans} based approach (right) shows higher mean $ACC$ over all embeddings.
Generally, \texttt{KMeans} plots show higher variance in the $ACC$ results.
The high variance at the AdaBoost model (yellow) for small $k$ ($<25$) is caused by some outliers in the $ACC$ results.
When comparing the different ML methods, AdaBoost generally performs best (yellow).
Comparing the embedding based results with the baseline generalizations, for $k<100$, the embedding labeled plots generally have higher $ACC$.
When comparing to un-anonymized $ACC$ obtained by the different ML methods, only few embedding-labeled $ACC$ scores obtained by AdaBoost have higher scores for results at $k\leq 50$, meaning
the $ACC$ for anonymized data is generally lower than the non-anonymized baseline but remains competitive and can even outperform.
Anonymized data show a noticeable decline as $k$ increases.

\begin{figure*}[h!]
    \centering
    \begin{subfigure}{0.49\linewidth}
        \includegraphics[width=\linewidth]{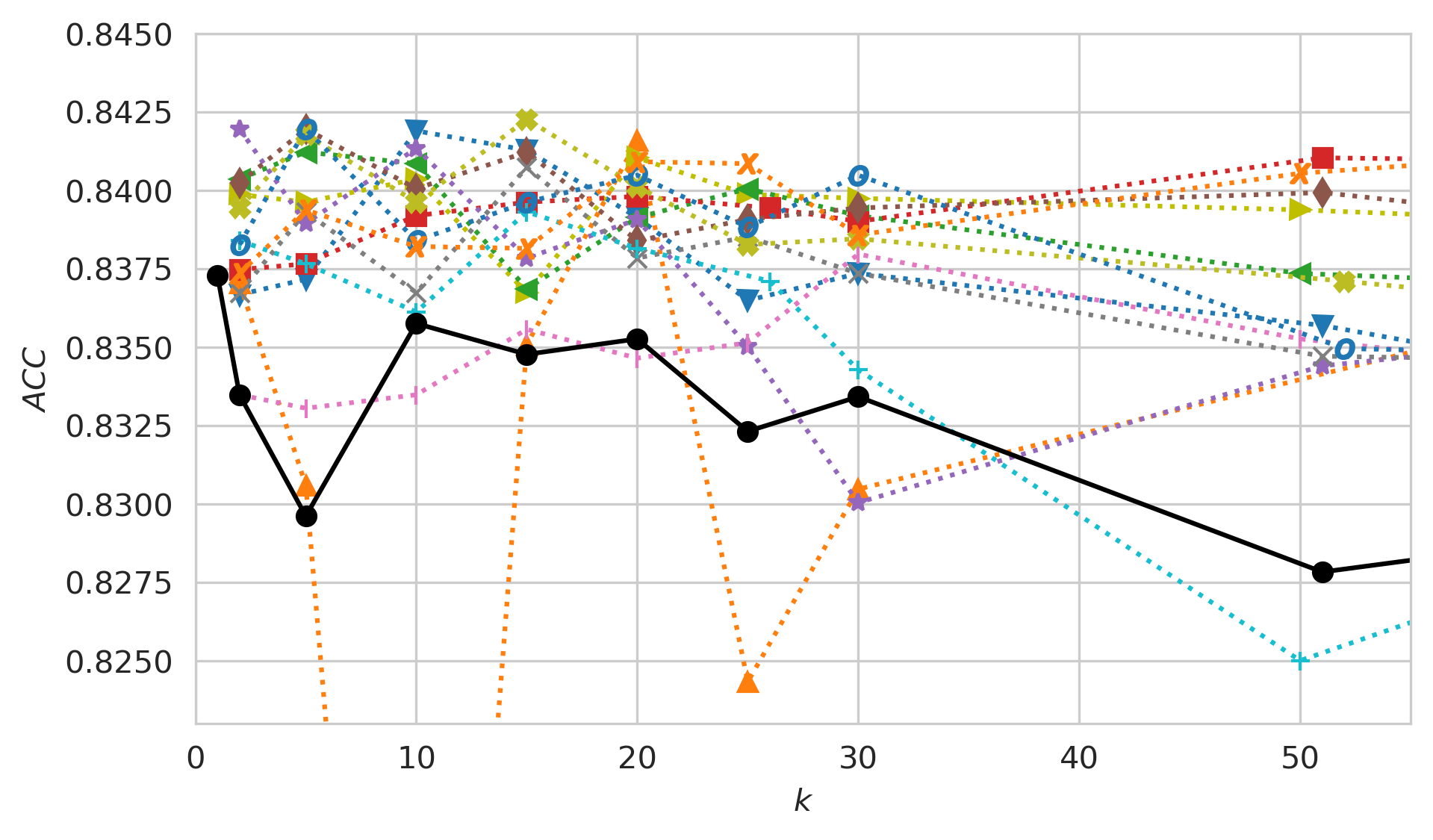}
        \caption{$ACC$ (AdaBoost)}
        \label{fig:subfig2}
    \end{subfigure}
    \hfill
    \begin{subfigure}{0.49\linewidth}
        \includegraphics[width=\linewidth]{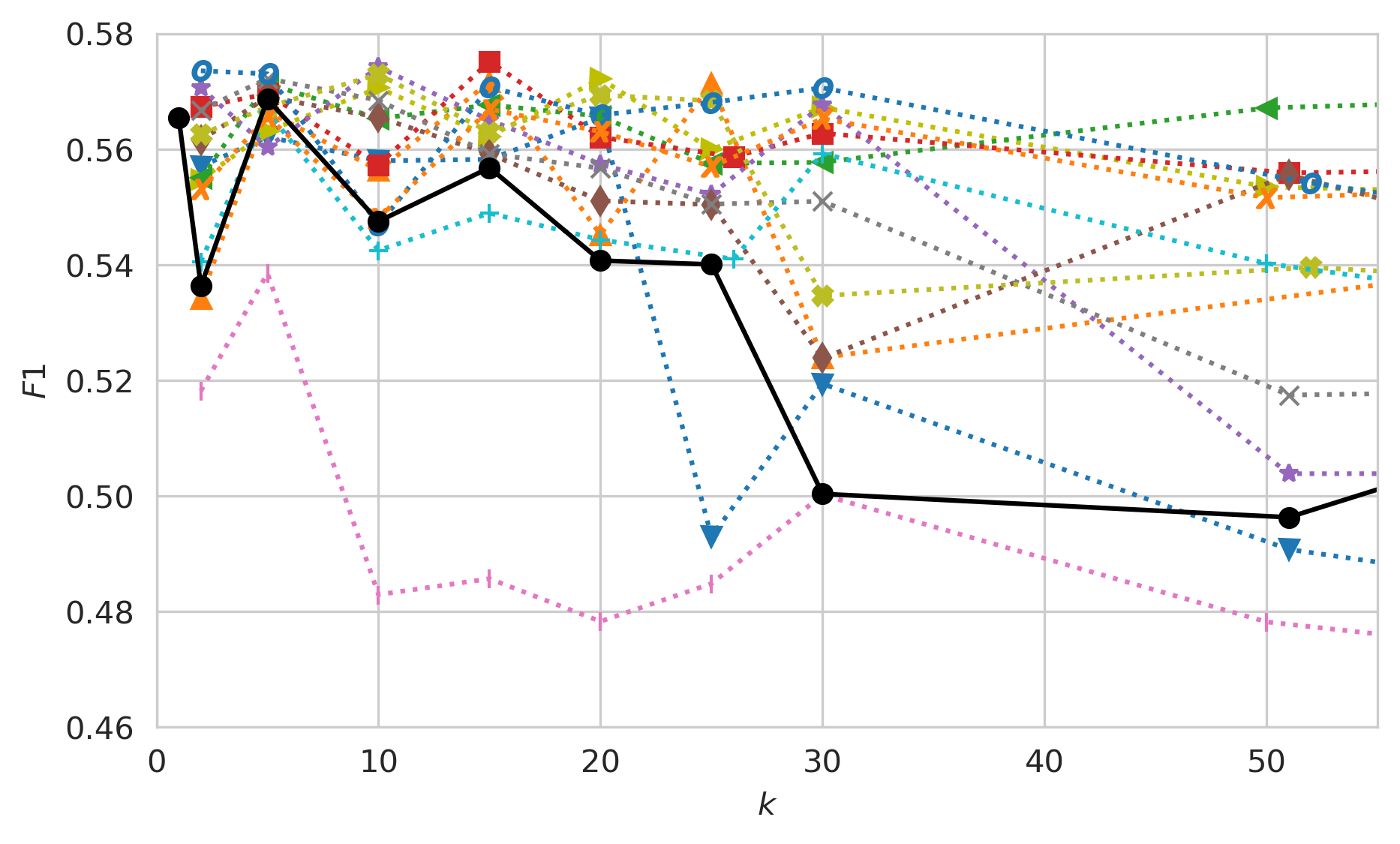}
        \caption{$F1$ (AdaBoost)}
        \label{fig:subfig3}
    \end{subfigure}
    \vskip 1mm
    \begin{subfigure}{0.49\linewidth}
        \includegraphics[width=\linewidth]{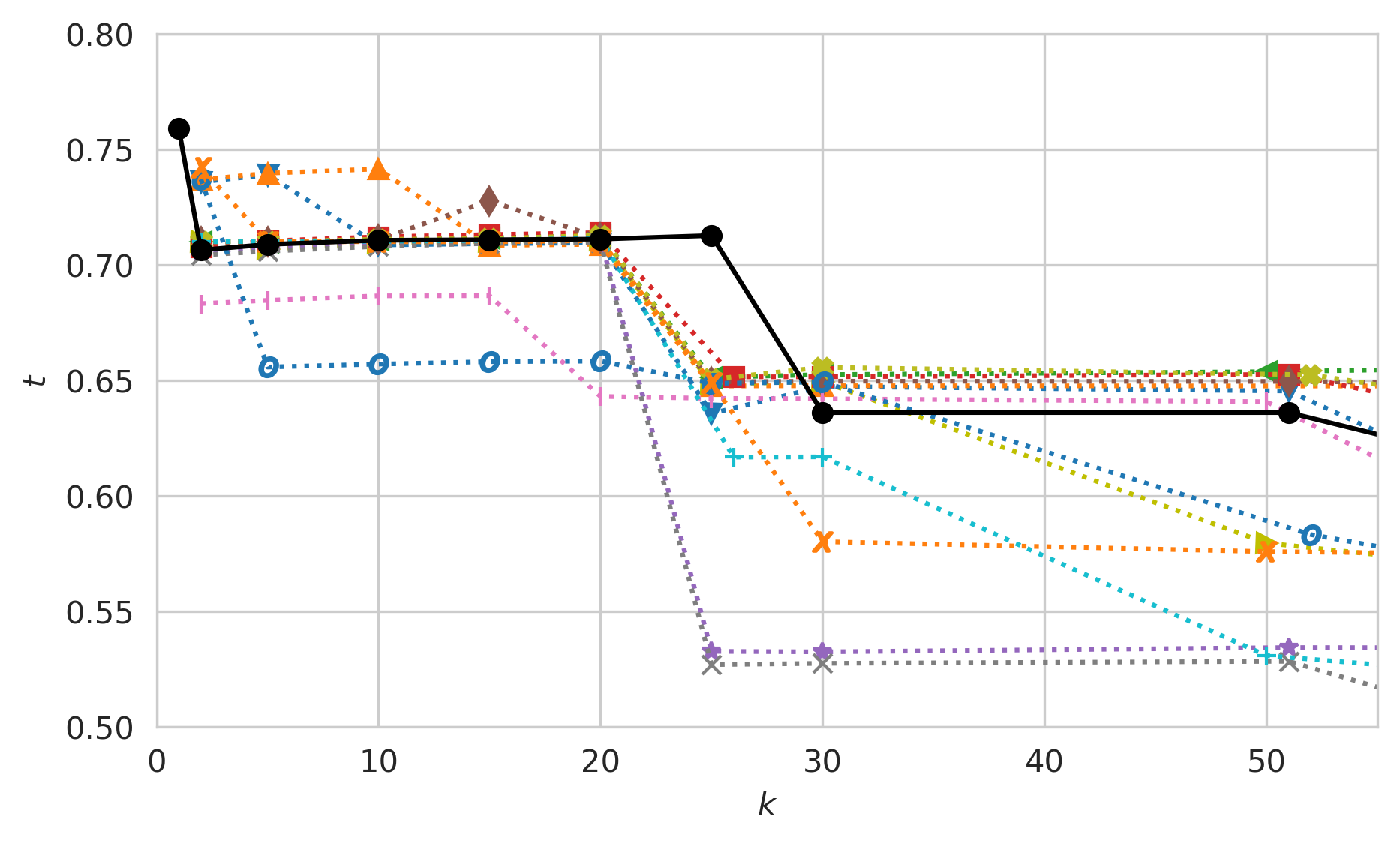}
        \caption{$t$-closeness}
        \label{fig:subfig1}
    \end{subfigure}
    \hfill
    \begin{subfigure}{0.49\linewidth}
        \includegraphics[width=\linewidth]{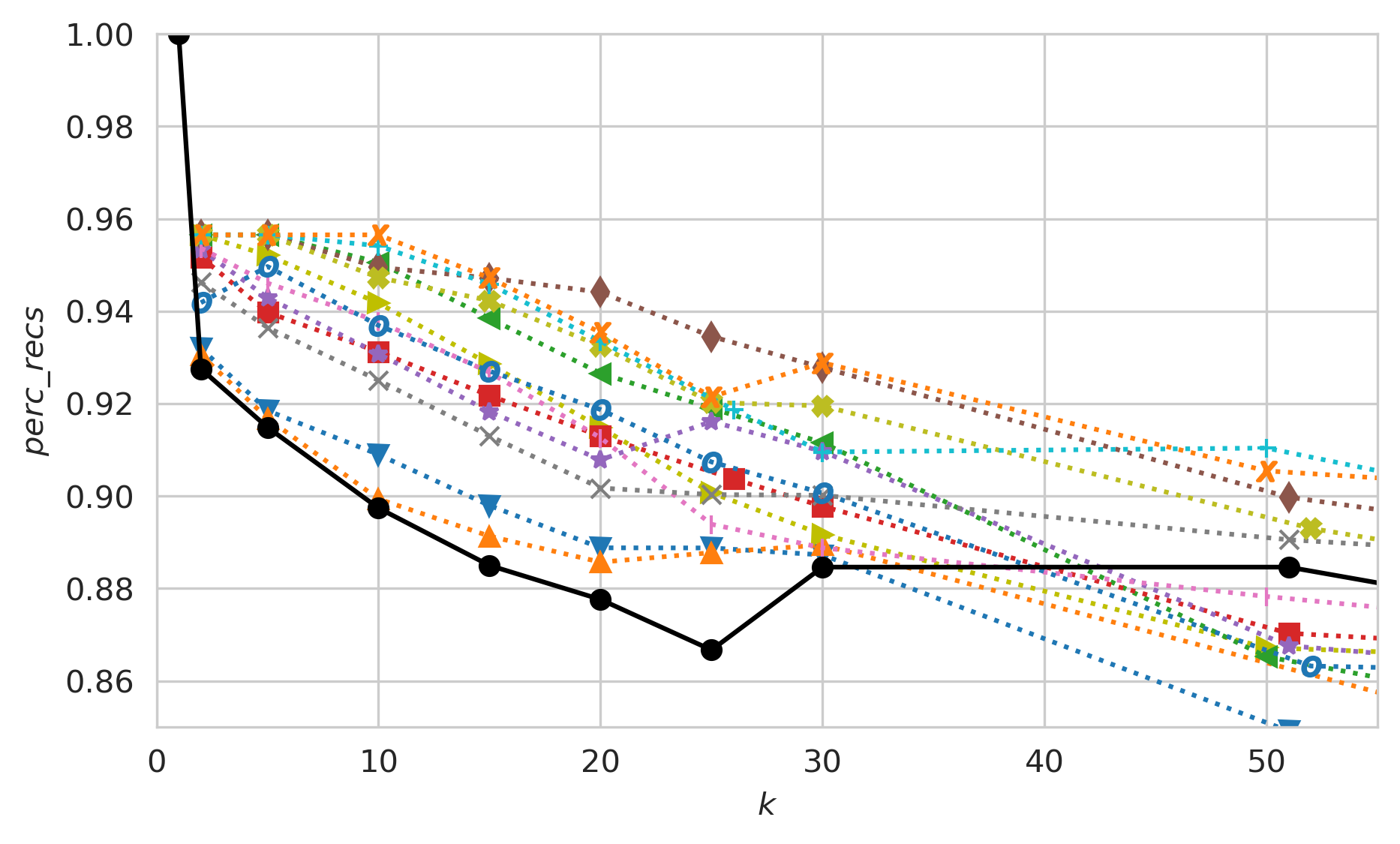}
        \caption{$perc\_recs$}
        \label{fig:subfig4}
    \end{subfigure}
    \vskip 1mm
    \begin{subfigure}{0.49\linewidth}
        \includegraphics[width=\linewidth]{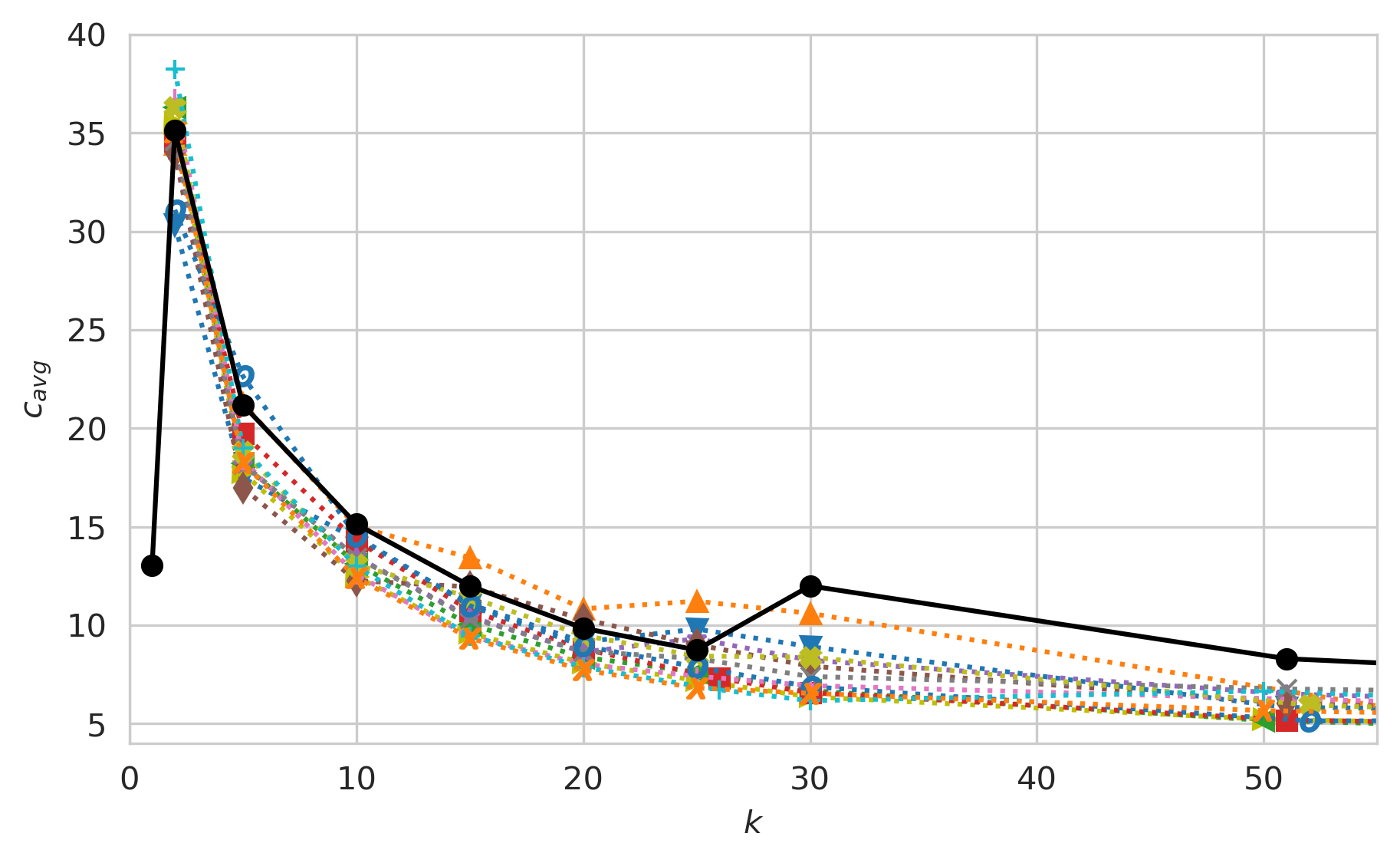}
        \caption{$c_{avg}$}
        \label{fig:subfig5}
    \end{subfigure}
    \hfill
        \begin{subfigure}{0.35\linewidth}
        \includegraphics[width=\linewidth]{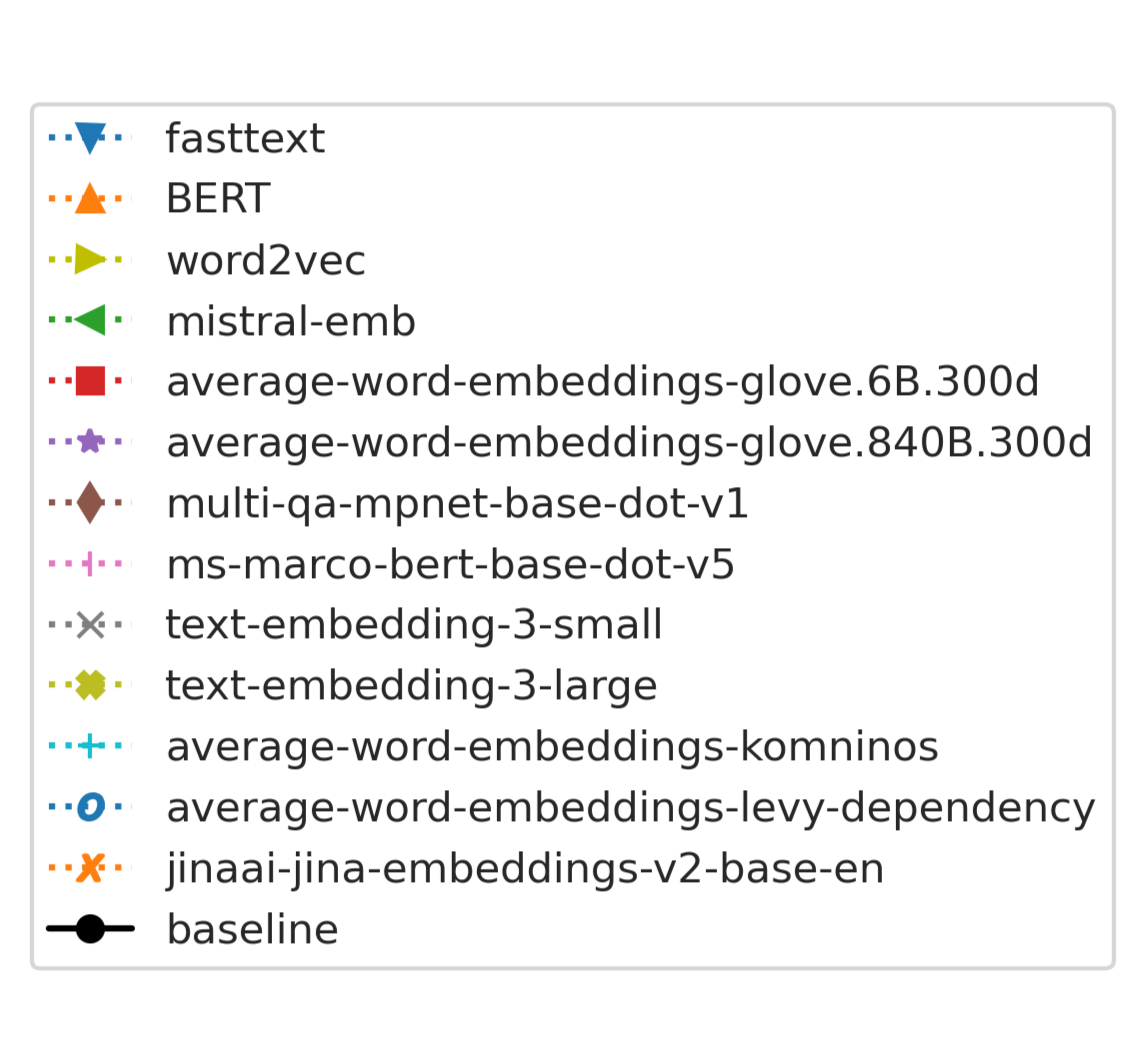}
        \caption{legend}
        \label{fig:subfig6}
    \end{subfigure}
    \caption{Performance measures. The VGHs used for anonymization were generated using \texttt{KMeans} clustering.}
    \label{fig:overall}
\end{figure*}

Fig. \ref{fig:overall} shows privacy, utility, and ML efficacy when using iterative \texttt{KMeans} clustering to obtain the VGHs. 
Lower $k$ values ($2$, $5$, $10$) tend to show better performance, with some variations depending on the specific embedding.
In comparison to \cite{Diaz2023}, $ACC$ behaves similar with growing $k$. 
We only plot $k \leq 50$ due to its higher practical relevance compared to greater $k$ values. Note, that obtained $k$-anonymity sometimes differs to input $k$ in the anonymization algorithm.
In Subfig. \ref{fig:subfig1}, the plots show a declining trend in $t$-closeness as $k$ increases, which is expected since higher $k$ typically means more generalization and potential information loss. 
Comparing embedding labels with the baseline label, no clear tendency can be seen. We assume that generalizing semantic similar values has stabilizing impact on sensitive attribute distributions in groups. Only the growing group sizes lead to similar group and global distributions. While increasing $k$ impacts baseline $ACC$ negatively, certain embeddings can maintain or even improve $ACC$ (Subfig. \ref{fig:subfig2}). 
$k$ does not directly correlate with $ACC$ -- especially for small $k$. We assume, that generalization and suppression to obtain decent $k$ might positively influence accuracy by removing outliers and promoting better model generalization capability when training. Except for outliers (\emph{BERT}) at $k=10$ and $k=25$, the anonymized data labeled by embeddings mostly yields higher $ACC$ up to $k=30$. The baseline starts high (with no anonymization applied) and decreases steeply, then slightly recovers at $k=30$. Other settings follow a similar trend but maintain higher $perc\_recs$ compared to the baseline as $k$ increases (Subfig. \ref{fig:subfig4}). From $k=5$, other settings generally show better 
$c_{avg}$ values compared to the baseline, especially for $2 \leq k \leq 30$ (Subfig. \ref{fig:subfig5}). Smaller values indicate better performance in anonymization as group sizes are closer to $k$. If all groups in the anonymized data have group size $k$, $c_{avg}=1$ is optimal.

\section{Conclusions}
\label{conclusion}

In summary, this paper introduces an anonymization pipeline, \texttt{ClustEm4Ano} (dataflow in Fig. \ref{fig:overall_system}), that utilizes embeddings of nominal textual attribute values. The paper explores the use of generating VGHs from open and closed-source embeddings in the context of microdata anonymization. The VGHs are obtained through hierarchical clustering of the embeddings by iteratively applying \texttt{KMeans} or \texttt{Agglomerative Hierarchical Clustering}. In Section \ref{experiments}, it tests the effectiveness of clustering text embeddings to obtain VGHs for generalization and suppression-based anonymization.
We show that various embeddings can be used to find VGHs that seem to be semantically correct (cf. Fig. \ref{fig:vis-emb}). The experimental results obtained show that using automatically generated VGHs with many levels can yield anonymizations that outperform the use of manually constructed VGHs in anonymization and can even outperform un-anonymized data, especially for small $k$ ($2 \leq k\leq 30)$. 

Despite the progress made, several limitations and areas for future work remain. 
Current clustering methods might not be effective at finding clusters in embeddings.
While the number of clusters is predefined, the optimal number might differ. Also, our method primarily focuses on the specific privacy models $k$-anonymity and $l$-diversity and the anonymization algorithm \texttt{FLASH} \cite{Kohlmayer2012}, leaving room for exploration with other privacy models and other anonymization algorithms.
Furthermore, in future research, domain-specific embeddings present an interesting area for exploration, as the choice of embedding models could be tailored to individual domains. Fine-tuning transformer-based word embeddings based on specific domains holds promise. However, in this regard, it is important to note that sharing word embeddings trained on sensitive data can compromise privacy, c.f. \cite{Abdalla2020a}.

\begin{credits}
\subsubsection{\ackname} This research work results from the research projects EAsyAnon (``Empfehlungs- und Auditsystem zur Anonymisierung.'') - funded by the German Federal Ministry of Education and Research (BMBF) and the European Union (grant number 16KISA128K) - and jbDATA - funded by the German Federal Ministry for Economic Affairs and Climate Action of Germany (BMWK) and the European Union (grant number 19A23003H).

\end{credits}

%

\bibliographystyle{splncs04}
\bibliography{paper}

\end{document}